\documentclass[11pt]{article} 
\usepackage{rldmsubmit,palatino}
\usepackage{graphicx, wrapfig}
\usepackage{amsmath}
\usepackage{subcaption}

\title{Autonomous state-space segmentation \\for Deep-RL sparse reward scenarios}

\author{
Gianluca~Maselli\textsuperscript{1,2}, Vieri Giuliano\textsuperscript{1} Santucci \\
\textsuperscript{1}Institute of Cognitive Sciences and Technologies (ISTC)\\
National Council of Research (CNR), Roma, Italy\\
\textsuperscript{2}  Department of Computer, Control and Management Engineering (DIAG)\\
Sapienza Univerity of Rome\\
\texttt{\{gianluca.maselli, vieri.santucci\}@istc.cnr.it} \\
}

\begin{document}

\maketitle

\begin{abstract}



Dealing with environments with sparse rewards has always been crucial for systems developed to operate in autonomous open-ended learning settings. Intrinsic Motivations could be an effective way to help Deep Reinforcement Learning algorithms learn in such scenarios. In fact, intrinsic reward signals, such as novelty or curiosity, are generally adopted to improve exploration when extrinsic rewards are delayed or absent. Building on previous works, we tackle the problem of learning policies in the presence of sparse rewards by proposing a two-level architecture that alternates an `ìntrinsically driven'' phase of exploration and autonomous sub-goal generation, to a phase of sparse reward, goal-directed policy learning. The idea is to build several small networks, each one specialized on a particular sub-path, and use them as starting points for future exploration without the need to further explore from scratch previously learnt paths. Two versions of the system have been trained and tested in the Gym SuperMarioBros environment without considering any additional extrinsic reward. The results show the validity of our approach and the importance of autonomously segment the environment to generate an efficient path towards the final goal.
\end{abstract}


\acknowledgements{This work was funded by  the European Union’s Horizon 2020, Research and Innovation Programme, GA 101070381 (“PILLAR-Robots -
Purposeful Intrinsically-motivated Lifelong Learning Autonomous Robots”)
\\
\\
This work has been accepted at the Multi-disciplinary Conference on Reinforcement Learning and Decision Making (RLDM), June 11-14, 2025. Dublin, Ireland.}

\startmain 

\section{Introduction}
Exploring vast state-action spaces in Reinforcement Learning (RL) presents significant challenges. Despite the impressive performances made through the integration of Deep Learning techniques \cite{mnih2015human}, in sparse or delayed reward scenarios agents often struggle to discover optimal policies due to limited exploration and inefficiencies in propagating reward signals, even with advanced techniques like Prioritized Experience Replay \cite{schaul2015prioritized}.  Addressing these challenges is crucial for applying artificial systems in scenarios where neither the agent nor the designer has prior knowledge of the world or the tasks to be accomplished. This aligns with the aims of the broader framework of Open-Ended Learning (OEL) \cite{sigaud2023definition}, where systems leverage self-generated signals to autonomously acquire competencies and knowledge \cite{colas2021intrinsically}.

Building on the concept of OEL, and particularly on the framework of Intrinsic Motivations (IMs), recent approaches have emerged to foster exploration in environments without dense rewards \cite{bellemare2016unifying, pathak2017curiosity, pathak2019self, eysenbach2018diversity}. Most of these approaches use curiosity as a novelty-based intrinsic reward signal to enable state-space exploration and discover extrinsic rewards more effectively than systems without self-generated signals. While further advancements have been made with model-based approaches \cite{sekar2020planning, mendonca2021discovering}, a significant limitation remains: most existing solutions require dense rewards to learn robust policies across the entire state space. Novelty-based strategies, such as the Intrinsic Curiosity Module (ICM) \cite{pathak2017curiosity}, help exploration by rewarding agents for discovering new states. However, these methods may still fall short in large environments, where agents risk frequently revisiting previously explored areas during trials or episodes. This not only wastes resources but also risks degrading the exploration policy itself, necessitating specific solutions to address this issue.
A different approach, following ideas form hierarchical RL \cite{pateria2021end}, is to split the main task into sub-goals. Sub-goal segmentation allows the state space to be divided into manageable parts, where the agent can learn specific policies to achieve intermediate goals. However, creating effective sub-goals often requires prior knowledge of the task, which, if provided by design, closely mirrors the use of dense reward functions, thus reintroducing the limitation we seek to avoid.

We propose here a two-level architecture that combines the exploration capabilities provided by the ICM module with the advantages of learning robust goal-directed policies according to autonomously generated sub-goals. Unlike previous models, our approach uses the ICM not only for exploration but also for autonomously segmenting the state space into meaningful sub-goals based on a general criterion. Once these sub-goals are discovered, the PTR learns specific policies for each sub-goal, that are used to improve both exploration efficiency (as starting point for further ICM-driven activity) and task performance in environments with sparse rewards. The concept of "competence", i.e. system ability to improve its performance by refining its own skills  \cite{santucci2013best}, is central to this approach, ensuring that the agent focuses only on those generated sub-goals where it can actually form a suitable policy, thus avoiding the risk of trying to achieve positions of the state space that have been achieved only by chance while exploring. We validate this architecture in the Gym SuperMarioBros environment\footnote{The repository can be found at https://github.com/Kautenja/gym-super-mario-bros}, where dense rewards have been explicitly disabled, showing that it enables efficient exploration and robust learning without relying on prior task-specific knowledge.

\section{Implemented solution and experimental scenario}

\begin{wrapfigure}{r}{3.2cm}
\centering
\includegraphics[width=3cm]{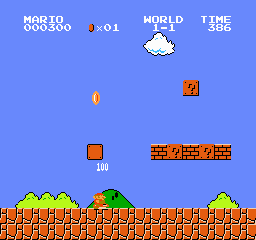}
\caption{The experimental scenario}
\label{fig:1}
\end{wrapfigure}
To preliminarily test our proposal, we chose the experimental scenario used in the work where the ICM model was presented as a way to integrate intrinsic exploration with dense rewards functions \cite{pathak2017curiosity}. Despite being an environment with peculiar characteristics (a side-scrolling game), SuperMarioBros presents a large state-and-actions space that is very complex to deal with when all dense rewards are excluded. In particular, we run our experiments on the first level of the game (Fig.~\ref{fig:1}).

\begin{wrapfigure}{r}{6.5cm}
\centering
\includegraphics[width=6.5cm]{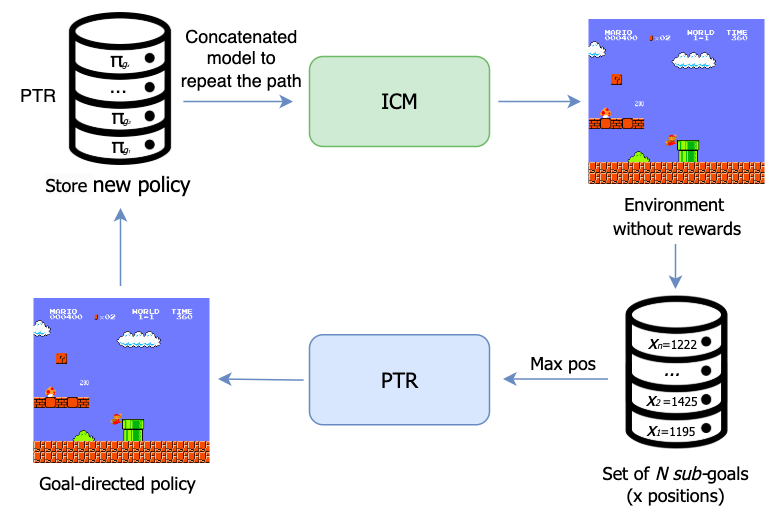}
\caption{The proposed architecture.}
\label{fig:arch}
\end{wrapfigure}

Our system is implemented as a two-components architecture (Fig.~\ref{fig:arch}): the exploration is performed by the ICM module, while the (sub-)goal directed policies are learnt by the PTR component. The ICM module receives as input the last concatenated 4 frames~\cite{mnih2015human} and is responsible for generating a novelty signal (determined by the error in predicting future encoded states) which is used to reinforce an explorative policy that determines the actions of the artificial agent \footnote{Where not explicitly mentioned, our implementation of the ICM module closely follow the original one that can be found in \cite{pathak2017curiosity}}. Differently from the original work we used a Double DQN \cite{van2016deep} to implement the explorative policy. The exploration phase runs for $N$ episodes ($N = 10$ in these experiments), each one ending if the agent ``dies'' or after 4,000 actions\footnote{The maximum number of steps within an episode is 24,000 but as in the original paper we use a frame skip of 6 during which the agent is repeating the same action before chosing a new one}, in each of which candidate sub-goals $g$ are collected. Due to the peculiar nature of the SuperMarioBros environment, we used a gamer-like perspective as a general criterion to identify candidate sub-goals: what is saved is the last state (as $x$ coordinate) reached in after an ICM-guided run, assuming that in this type of games what a player would set as a sub-goal is to be able to systematically reach the last ``known'' state, and then learn new strategies to advance from it. Different experimental environments may of course need different criteria: in a robotic environment it might for example make sense to use the changes generated in the world as target states, but identifying the best criteria is not the purpose of this paper.

Once candidate $g_x$ are collected, the PTR will use the one with highest $x$ (the furthest) to train a (sub-)goal-specific policy $\pi^g_x$ thanks to the sparse pseudo-reward for achieving $g_x$. In particular, the reward $r^{g_x}$ used to train $\pi^g_x$ is
\begin{equation}
    r^{g_x} = \begin{cases}
      1~\text{if}~ x_t \ge g_x \\
      0~\text{elsewhere}
    \end{cases}\,.
\end{equation}
meaning the agent receives the reward if and only if it reaches the goal position. 
In this way we keep the sparseness of the environment, but the agent ``only'' has to learn shorter policies in smaller subsets of the entire state space $S^{g_x} \subset S$. To regulate the trade-off between exploration and exploitation, to train the sub-goal policies we implemented a Noisy DQN\footnote{While receiving as input the same amount of stacked frames, this network uses 4 skipping frames for action repetition} \cite{fortunato1706noisy} trained with a Prioritized Experience Replay buffer~\cite{schaul2015prioritized}. 
However, some $g_x$ may have been discovered through a behaviour which, by chance, managed to get to distant positions that were difficult to reach, thus effectively reintroducing the problem of learning by sparse rewards even the shorter policies $\pi_{g_x}$. For this reason we leverage another concept from intrinsically motivated open-ended learning, i.e. the one of ``competence''. In particular, the competence $C$ is as a moving average of the rewards obtained for achieving $g_x$ in the the last $K=30$ learning episodes  

\begin{equation}
    C_{g_x} = \frac{1}{K} \sum_{i=0}^{K-1} r_{j-i}
    \label{eq3}
\end{equation}

where $j$ is the current training episode within a maximum of $J=10,000$. If within $J$ the agent is not able to achieve a target level of competence (here set at $0.9$), the PTR will move to learn the second most distant $g_x$. Differently, if the level is achieved a test phase is performed, and if the target sub-goal is achieved the related policy $\pi_{g_x}$ is stored.

At this stage, a new exploration phase begins. Unlike standard approaches that restart exploration from the initial state (or from a random state), our method, building on ideas such as GO-EXPLORE \cite{ecoffet2019go} and extending them further, leverages the agent’s previously acquired knowledge to navigate to the farthest point in the environment where it has demonstrated high competence. In particular, all previously learned policies from the PTR are executed sequentially to efficiently reach the last sub-goal state, from which exploration resumes under the guidance of the ICM module. Obviously, a scenario such as SuperMarioBros facilitates this approach, since in a side-scrolling game the final state achieved through one policy effectively becomes the starting point of the next one, and the linear structure of the environment ensures that no ``branching trajectories'' occur. Choosing a strategy to determine which policies to replay for resuming exploration may require task-specific adjustments depending on the environment. However, leveraging intrinsic dynamics can lead to more general solutions, as demonstrated even in real-world robotics environments (as shown in \cite{Romero2024GoalDiscovery}).

\subsection{Compared systems}

The comparative baseline for these preliminary experiments is provided by the original work introducing the ICM module \cite{pathak2017curiosity}, where the authors showed that guided solely by the novelty signal from the ICM, the agent was able to reach only 38\% of the first level (around position $x=1220$). Notably, we also run an experiment where, in addition to the ICM signal, we provide a sparse reward for achieving the final state of the level: the agent was able to reach it (on average) just once in 50,0000 training episodes, which is not sufficient to allow for the creation of a policy in the testing phase. This result shows how in such an environment IMs alone are not sufficient to allow the proper exploration of the state space.

Based on these premises, we present data from two systems: one is the system described in Sec. 2, and the other is a modified version in which the only difference is that the exploratory phase is run for a single episode, instead of 10 as in our proposal. In this version, PTR learns a policy to reach a single identified sub-goal $g_x$. If the agent fails to reach $g_x$ within the set time, no policy $\pi_{g_x}$ is stored, $g_x$ is discarded, and a new $g_{x'}$ is identified through another exploratory phase. This modified system was tested to evaluate the importance to autonomously determine how to partition the state-space. Stopping at the first identified sub-goal may result in excessively small (sub-)policies, or wasted time trying to reach complex positions with sparse rewards. Conversely, using a larger set of possible target sub-goals, starting with more complex ones and eventually moving to simpler ones, should enable the system to autonomously find the optimal granularity for partitioning the environment, depending on the specific area it is interacting with.


\section{Results and Discussion}

While systems relying solely on intrinsic motivations and sparse rewards failed to reach the final state of the environment (position $x=3161$), both versions of our system successfully created a sequence of sub-paths which, when linked together, enabled the agent to achieve the final goal.

\begin{wrapfigure}{r}{8cm}
\centering
\includegraphics[width=8cm]{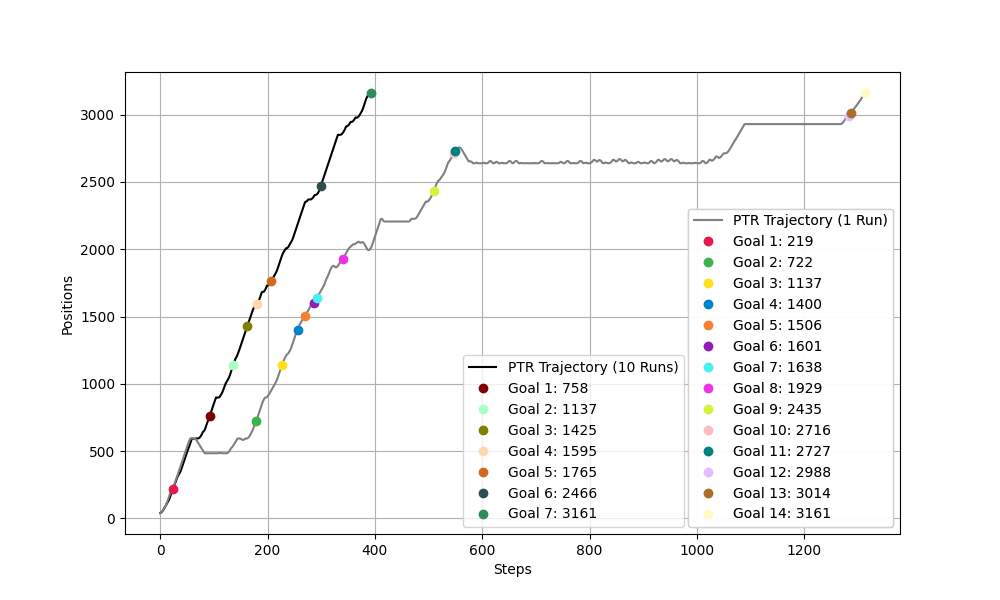}
\caption{The paths generated by the compared systems of one representative seed}
\label{figure:c}
\end{wrapfigure}

Fig. \ref{figure:c} shows the paths generated by one representative seed reflecting the typical behaviour of the two systems during the experiments. Indeed, each experiment run with a different seed resulting in different trajectories for both the systems described in Sec. 2. For these reasons we chose to show the results of only one seed reflecting the average performance of the systems.

These results confirm that allowing the system to autonomously partition the state space into sub-goals, learn specific policies to reach them, and then leverage these policies to drive further exploration, enables the agent to solve a task that would otherwise be extremely complex. However, both when examining individual seeds and analyzing the average results (Table 1, based on 10 seeds for each system), it is evident that the two methods used to generate sub-goals lead to significant differences in the performance of the two systems. 

\begin{table}
\centering
\renewcommand{\arraystretch}{1.5} 
\setlength{\tabcolsep}{12pt}      
\begin{tabular}{|p{6em}|c|c|c|}
\hline
\multicolumn{4}{|c|}{\textbf{Average Results}} \\ \hline
& Sub-goals & Episodes & Steps \\ \hline
ICM $N=10$ & \textbf{6.8} ($\sigma$=1.6) & 78,985.4 ($\sigma$=21,020) & \textbf{375.6} ($\sigma$=83.5) \\ 
ICM $N=1$  & 14.6 ($\sigma$=2.06) & \textbf{41,344.4} ($\sigma$=14,374) & 837.4 ($\sigma$=375.7)\\ \hline
\end{tabular}
\caption{Averages results and standard deviations (over 10 replications of the two versions of the proposed systems) of the number of created sub-goals, episodes to reach the final goal, and steps needed by the chained sub-policies to reach the final goal.}
\label{table:1}
\end{table}

\begin{figure}[h!]
\centering
\begin{subfigure}{.4\textwidth}
    \centering
    \includegraphics[width=.95\linewidth]{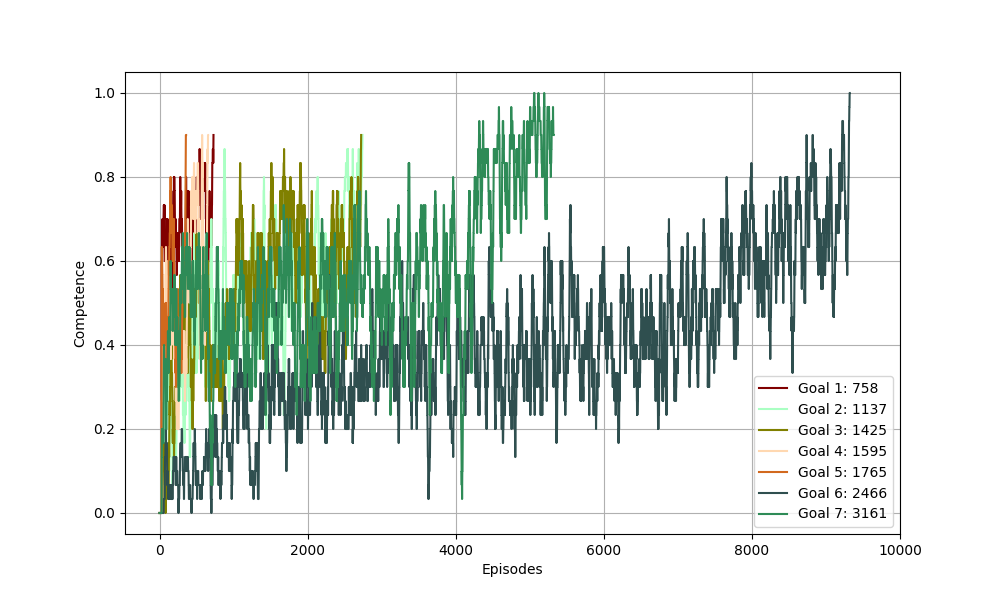}  
    \caption{ICM $N=10$}
    \label{figure:a}
\end{subfigure}
\begin{subfigure}{.4\textwidth}
    \centering
    \includegraphics[width=.95\linewidth]{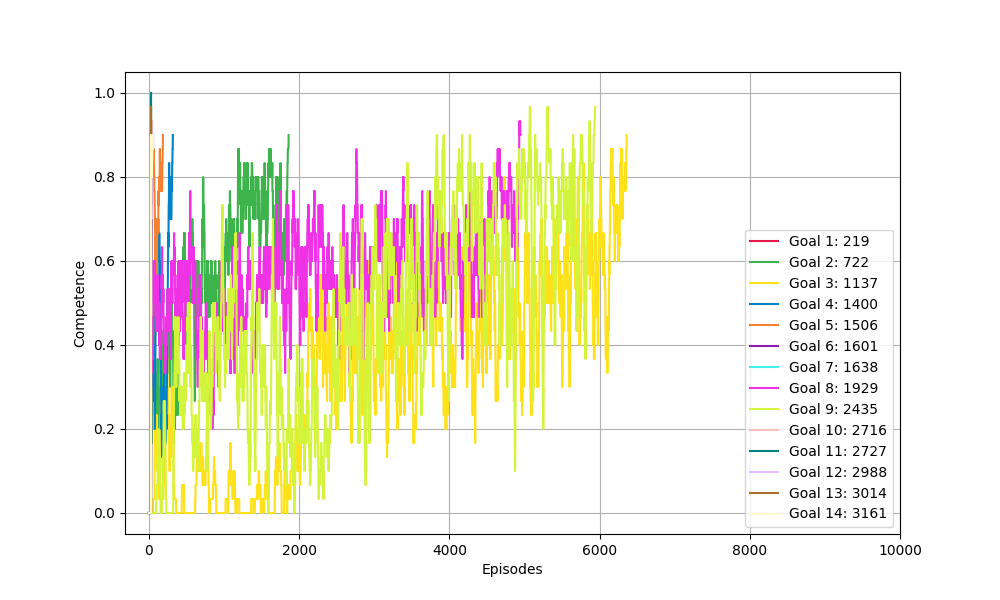}  
    \caption{ICM $N=1$}
    \label{figure:b}
\end{subfigure}
\caption{Competence curves (over number of training episodes) over the discovered sub-goals in the two compared systems. Data refers to the same representative seeds used in Fig.3}
\label{Fig:Competence}
\end{figure}

The system that generates candidate sub-goals based on 10 exploratory episodes guided by the ICM is able to produce more robust and direct sub-policies. These, when combined, allow the agent to reach the end of the level with a significantly reduced number of steps (i.e., actions) compared to the system that relies on single exploratory episodes for sub-goal discovery. This difference arises from the fact that the system which takes the first position identified in an ICM-guided run as the new sub-goal, frequently finds positions very close to the starting point of the exploration. Such positions are easily reachable, even by policies that, not yet fully optimized, perform almost random behavior. The result is a set of sub-policies that are generally shorter, but which, in relative terms, require more time to execute, since the system assumes it has achieved adequate competence (based on reaching the target state), even though this competence is derived from a sequence of inefficient actions. This is evident from the results showing the competence level over time for the sub-goals generated by the representative seeds of the two systems (Fig. \ref{Fig:Competence}, same seeds as in Fig. \ref{figure:c}). The system that uses 1 ICM-guided episode generates sub-goals for which it can reach the necessary competence much more quickly, allowing it to proceed to the next exploratory phase. However, when we combine this with the fact that the average number of steps required by this system to complete the main task is higher, we can see that these policies are, in fact, less efficient than the (longer) ones generated by our system.

The total number of episodes used by the system with only one exploratory run at a time is, on average, lower than that of the version we proposed. This is because our system not only needs to perform a greater number of explorations each time, but also may spend learning episodes with the PTR on complex positions for which it is not be able to reach the minimum competence threshold within the maximum time available for learning each sub-policy. However, allowing the system to have multiple candidate sub-goals enables it to partition the state space into a smaller number of sub-policies. This is not only advantageous in terms of reducing the risk of creating a large (and as shown, eventually inefficient) number of policies: starting from more complex targets and, if necessary, moving to simpler ones allows the system to autonomously partition the environment into the optimal (or near-optimal) number of necessary skills. This is further confirmed by the fact that the variance in the number of sub-policies created across different seeds is minimal ($\sigma = 1.6$). This result shows that our system, through repeated exploration and a competence-based criterion, has autonomously segmented the state space in the most suitable way, depending on the complexity of the area in which it is located: smaller sub-policies reflect regions where the system must perform more specific behaviors (e.g., due to the presence of obstacles or enemies), while longer sub-policies are associated with paths where it is easier for the agent to progress through the level.

\bibliographystyle{IEEEtran}
\bibliography{Mybib.bib}

\end{document}